%% file: main.tex
\documentclass[runningheads]{llncs}
\usepackage{subfig}
\usepackage[bb=boondox]{mathalfa}
\usepackage{xr-hyper}
\usepackage{url}
\usepackage{hyperref}
\usepackage{graphicx}
\usepackage{amssymb}

\usepackage{amsmath}
\newcommand{\method}{GKD }
\newcommand{\fig}{Fig.}

\newcommand{\supfig}{Supplementary Fig.}
\newcommand{\suptab}{Supplementary Table}

\newcommand{\teacher}{$\mathcal{D}^{T}_{\theta}$}
\newcommand{\student}{$\mathcal{D}^{S}_{\phi}$}

\usepackage{color}

\makeatletter
\newcommand\notsotiny{\@setfontsize\notsotiny\@vipt\@viipt}

\makeatother

\begin{document}
\title{GKD: Semi-supervised Graph Knowledge Distillation for Graph-Independent Inference}
\titlerunning{GKD: Graph Knowledge Distillation}
\author{Mahsa Ghorbani\inst{1,2}\and
Mojtaba Bahrami \inst{1} \and
Anees Kazi \inst{2}\and
Mahdieh Soleymani Baghshah \inst{1} \and
Hamid R. Rabiee \inst{1} \and
Nassir Navab \inst{2,3}}
\authorrunning{Ghorbani et al.}
%
\institute{Sharif University of Technology, Tehran, Iran \and
Computer Aided Medical Procedures, Technical University of Munich, Germany \and
Computer Aided Medical Procedures, Johns Hopkins University, Baltimore, USA\\
}
\maketitle              
\begin{abstract}
The increased amount of multi-modal medical data has opened the opportunities to simultaneously process various modalities such as imaging and non-imaging data to gain a comprehensive insight into the disease prediction domain. Recent studies using Graph Convolutional Networks (GCNs) provide novel semi-supervised approaches for integrating heterogeneous modalities while investigating the patients' associations for disease prediction.
However, when the meta-data used for graph construction is not available at inference time (e.g., coming from a distinct population), the conventional methods exhibit poor performance.
To address this issue, we propose a novel semi-supervised approach named \method based on the knowledge distillation. We train a teacher component that employs the label-propagation algorithm besides a deep neural network to benefit from the graph and non-graph modalities only in the training phase. The teacher component embeds all the available information into the soft pseudo-labels.  The soft pseudo-labels are then used to train a deep student network for disease prediction of unseen test data for which the graph modality is unavailable. 
We perform our experiments on two public datasets for diagnosing Autism spectrum disorder, and Alzheimer's disease, along with a thorough analysis on synthetic multi-modal datasets. According to these experiments, \method outperforms the previous graph-based deep learning methods in terms of accuracy, AUC, and Macro F1.
\end{abstract}
\section{Introduction}
Acquisition and exploitation of the wide range of genetic, phenotypic, and behavioral information along with the imaging data increase the necessity of an integrated framework for analyzing medical imaging and non-imaging modalities \cite{cai2019survey,abrol2019multimodal,huang2020fusion,bi2019effective}. 
On the other hand, with the success of conventional deep neural networks, there has been an extensive effort on utilizing them for multi-modal datasets \cite{guo2019deep,lee2019predicting,venugopalan2021multimodal}.
Considering the relationships between the patients based on one or multiple important modalities is beneficial as it helps to analyze and study the similar cohort of patients together. 
By viewing the patients as nodes and their associations as edges, graphs provide a natural way of representing the interactions among a population \cite{liu2020identification,yang2019interpretable}.
Meanwhile, Graph Convolutional Networks (GCNs) provide deep network architectures for integrating node features and graph modalities of data and recently have been used for such population-level analysis \cite{parisot2017spectral}.
In the non-medical domain, ChebyNet \cite{defferrard2016convolutional} was one of the first methods which extended the theory of graph signal processing into deep learning. Further, Kipf et al. \cite{kipf2016semi} simplified the ChebyNet by introducing a method named GCN to perform semi-supervised classification on citation datasets. Parisot et al. \cite{parisot2017spectral} exploited GCN for the disease prediction problem with multi-modal datasets. Then, GCN is adopted in different forms for medical applications including disease classification \cite{huang2020edge,kazi2019inceptiongcn,ghorbani2021ragcn}, brain analysis \cite{li2020braingnn}, and mammogram analysis \cite{du2019zoom}.

In practice, all the graph modalities are not always available for new or unseen patients. This can be due to missing observations or costly and time-consuming data collection for graph construction. Moreover, the test samples may not belong to the primary population with the same set of measured modalities.
In such a setting, where the new coming samples are isolated nodes with no edges connecting them to the rest of the population graph, the conventional graph-based models might fail to generalize to new isolated samples. 
This is due to the reason that GCN-based methods smooth each node's features by averaging with its neighbors in each layer of the network, which makes the features of adjacent nodes close to each other and dissimilar to the farther ones \cite{zhang2019graph}. 
This also makes it hard for the model, learned over the connected components, to perform as well on isolated or low-degree nodes with unsmoothed high-variance features.
To address this limitation, a comprehensive approach is required to make the most use of the extra modality information among the training samples.

In order to overcome this challenge, inspired by the knowledge distillation technique (\cite{bucilu2006model,hinton2015distilling}), we propose a novel student-teacher method to leverage all modalities at training and operate independently from the graph at inference time. Our proposed method encapsulates all the graph-related data using Label-Propagation Algorithm (LPA) into pseudo-labels produced by a so-called Teacher model. The pseudo-labels are then used to train a network (Student) that learns the mapping from high variance and noisy original input space to the output without filtering the features throughout the network. By doing so, the student network makes predictions without needing the presence of the graph modality on the future unseen data. 
\section{Knowledge Distillation Framework}
Complex models or an ensemble of separately trained models can extract complicated structures from data but are not efficient for deployment. 
The knowledge of a trained model is saved in its parameters, and transferring the learned weights is not straightforward while changing the structure, form, and size of the target model. 
However, a model's knowledge can be viewed as a mapping from its input to the output.
In a classification task, the relative magnitude of class probabilities of a trained model's output is a rich source of information and represents the similarity between classes. This similarity measure gained by a teacher network can be used as a source of knowledge to train a smaller and more efficient network (student).
Caruana et al. \cite{bucilu2006model} followed by Hinton et al. \cite{hinton2015distilling} show that the student model generalizes in the same way as the teacher and is superior compared to being trained by hard targets from scratch. 
Inspired by this framework, we utilize all the available modalities
only during the learning procedure of the teacher and then transfer the obtained knowledge into a student that does not have any assumption regarding the graph's availability at inference time. 
\section{Methods}
\begin{figure}[!tb]
    \centering
    \includegraphics[width=\textwidth,height=\textheight,keepaspectratio]{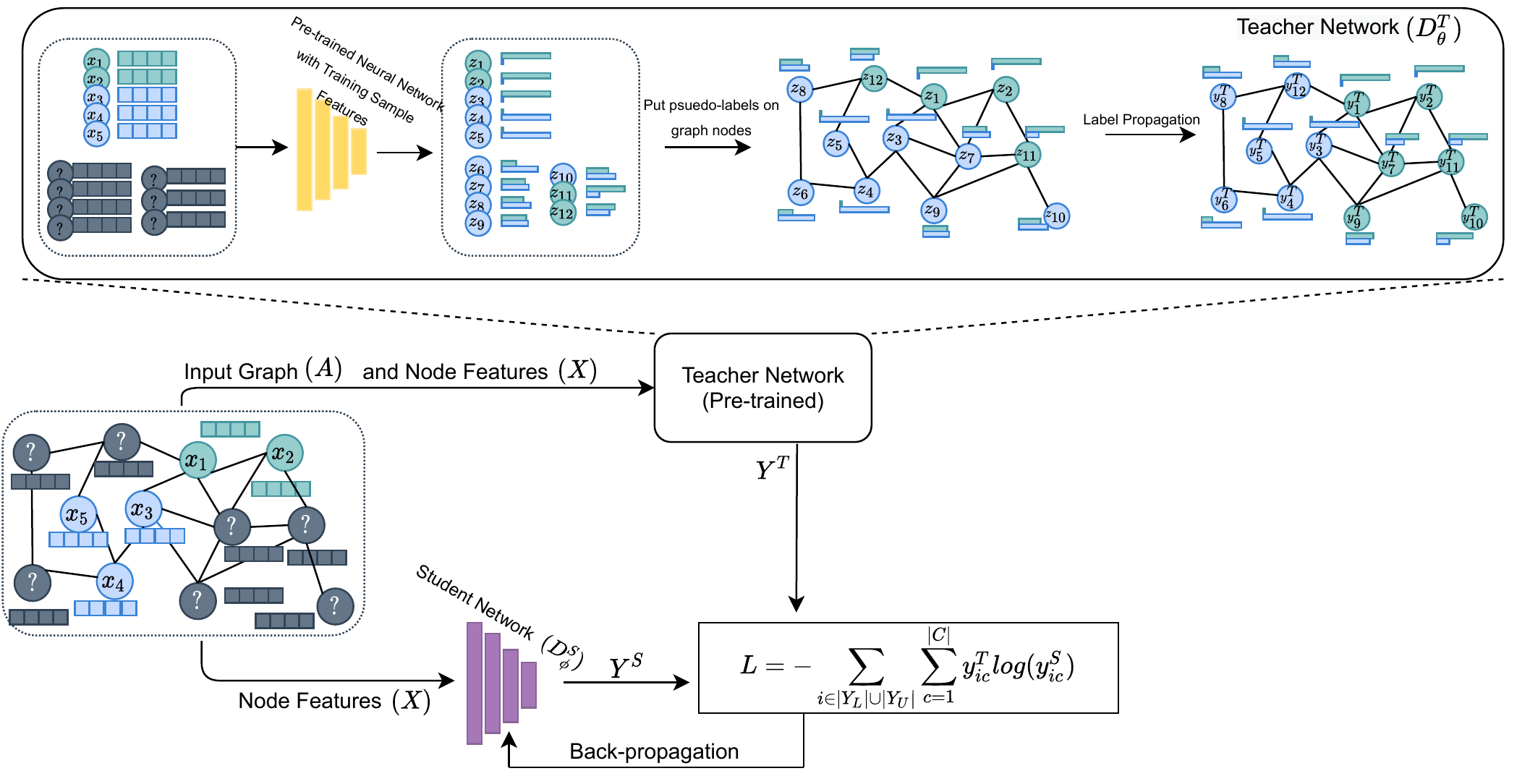}
    \caption{Overview of the proposed method. The training set comprises of labeled and unlabeled nodes. Unlabeled nodes are in dark grey and other colors depicts the node labels. The upper part: teacher network. The lower part: student network. The student is trained by cross-entropy loss between teacher and student outputs to inject the teachers knowledge into the student's parameters. }
    \label{fig:model}
\end{figure}

We provide a detailed description of our \method method by explaining the problem definition and notations, the teacher structure, and the student network that is the final classifier. An overview of the method is available in \fig~\ref{fig:model}.\\
\textbf{Problem Definition and Notations:}
Assume that $N$ patients are given. We represent the subject interactions graph as $G(V, E, A)$, where $V$ is the set of patient vertices ($|V|=N$), $E \in (V \times V)$ is the set of edges and $A\in  \mathbb{R}^{N \times N}$ shows the adjacency matrix. Let $D$ be the diagonal degree matrix for $A$.
We also define the $F$-dimensional node features as $X \in \mathbb{R}^{N \times F}$.
The state of patients is described by the one-hot labels $Y \in \mathbb{R}^{N \times C}$ where $C$ denotes the total number of states (e.g., diseased or normal).
According to our semi-supervised setting, the training samples consist of two mutually exclusive sets of labeled samples ($V_L$) and unlabeled ones ($V_U$) where $V = V_L \cup V_U$.  The ground-truth labels $Y_L$ are only available for $V_L$. 
Finally, our objective is to train a teacher to predict the class probabilities for unlabeled training nodes ($V_U$) and use them together with labeled training nodes as $Y^T$ to train a student network with cross-entropy loss function and apply it on unseen test samples. \\
\textbf{Teacher for Knowledge Integration:}
Here, we develop a mechanism to attain a teacher using both the node features and graph interactions.
To this end, we first try to extract the available knowledge from the node features by training a deep neural network \teacher, where $\theta$ shows the network's parameters. The teacher network is trained by the labeled training pairs ($X_L, Y_L$). 
Afterward, the trained network is applied to the rest of the unlabeled data ($X_U$) and predicts the best possible soft pseudo-labels $\hat{Y}_U$ for them. 
Together with the ground-truth labels $Y_L$, we initialize the graph $G$ with $Y_L \cup \hat{Y}_U$ as node labels. Up to this point, each patient's node labels are predicted independently of their neighbors in the graph.
To add the graph information to the node labels, we employ the well-known LPA to distribute the label information all around the graph.
However, applying the original LPA might forget the primary predictions for each node and result in the over-emphasizing on the graph. We try to compensate for this issue by adding a remembrance term to LPA, which avoids forgetting the initial predictions. The $k$-th iteration of our modified LPA is as follow:
\begin{equation}
\begin{split}
Y^{T^{(k)}} &= (1-\alpha) D^{-1}A Y^{T^{(k-1)}}+\alpha \underbrace{Y^{T^{0}}}_{\substack{\text{Remembrance}\\ \text{term}}};  \;\; Y^{T^0} = Y_L \cup \hat{Y}_U, \\
Y_{L}^{T^{(k)}} &= Y_{L},
\end{split}
\label{eq:label-prop}
\end{equation}
where $Y^{T^0}$ is the initial predictions by the teacher network, and $Y^{T^k}$ denotes the set of labels at $k$-th iteration. At each iteration, the labels of training nodes are replaced with their true labels ($Y_L$). The output of the last iteration is known as teacher soft pseudo-labels ($Y^T$). The effect of graph neighborhood and the remembrance term is adjustable by the coefficient $\alpha$. \\
\textbf{Student as Final Classifier for Inference Time:}
As described in distillation framework \cite{hinton2015distilling}, the student network is trained to mimic the input-output mapping learned by the teacher network. 
Specifically, the student network is trained using the class-probabilities produced by the teacher ($Y^T$).
For this purpose, we use another neural network \student with parameters $\phi$ called student network and train its parameters with the cross-entropy loss function between  $Y^{S}$ and $Y^{T}$ in which $Y^S$ is the output predictions of the student network \student.
The intuition behind minimizing the cross-entropy between $Y^{S}$ and $Y^{T}$ is that it is equivalent to minimizing the KL-divergence distance between the distribution of the $Y^{S}$ and $Y^{T}$ random variables, enforcing the student network to imitate the predictions of the teacher as a rich model.
\section{Experiments and Results}
In this section, we perform experimental evaluations to explore the performance of the proposed method for disease prediction compared with state-of-the-art methods. \\
\textbf{Experimental Setup:}
We show our results on two multi-modal medical datasets and analyze the behavior of the methods.
Finally, a set of multi-modal synthetic datasets are created and further analysis on the stability and robustness of the proposed method are carried out on them. 
We choose EV-GCN\cite{huang2020edge}, and the introduced method by Parisot et al. \cite{parisot2018disease} as two graph-based approaches incorporating GCN architecture designed for multi-modal disease prediction, DNN-JFC \cite{xu2016multimodal} as a multi-modal method, fully connected deep neural network (DNN) and the introduced method by Abraham et al. \cite{abraham2017deriving} as unimodal baselines. 
Since DNN-JFC uses the features of all modalities for classification, at the inference time, when one modality (graph modality) is not available, mean imputation is employed to fill the values for this method.
All the results are obtained from $5$ different initialization seeds. Fully-connected neural network is the selected structure for both teacher and student in the experiments. For all methods, we examined networks with one, two and three hidden layers with a set of different number of units and checked the set of \{$5e-3,1e-2$\} for learning rate and \{$0.1,0.3,0.5$\} for dropout. Adam optimizer is utilized for training of the networks \cite{kingma2014adam}. The results are reported on the test set using the best configuration (selected based on the validation set) for each method per dataset. 
\subsection{Autism Spectrum Disorder Diagnosis on the ABIDE dataset}
\begin{figure}[!htb]
{
  \begin{center}
    \subfloat[ABIDE Accuracy]{\label{fig:abide-acc}\includegraphics[width=0.33\textwidth]{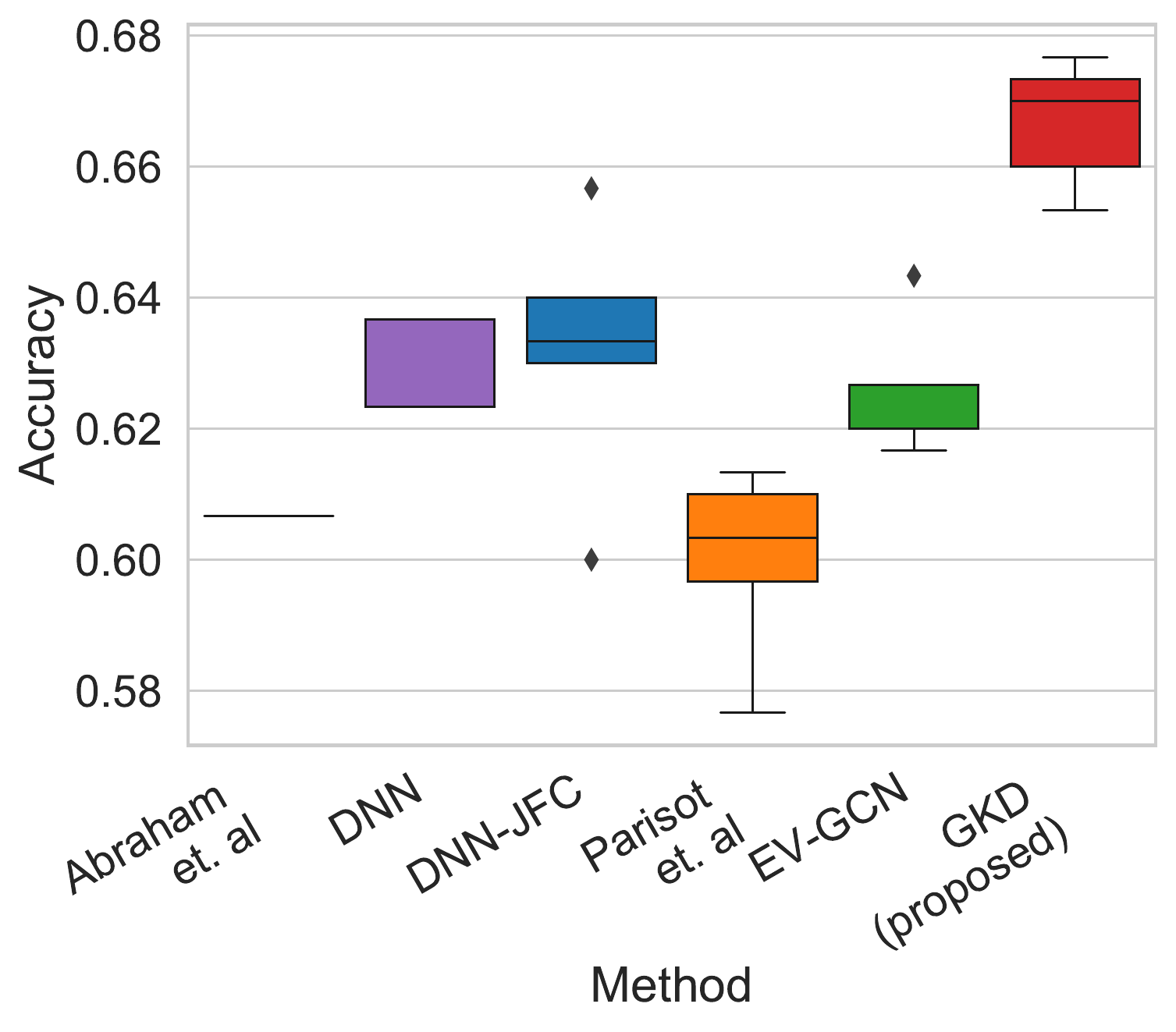}}
    \subfloat[ABIDE Macro F1]{\label{fig:abide-f1macro}\includegraphics[width=0.33\textwidth]{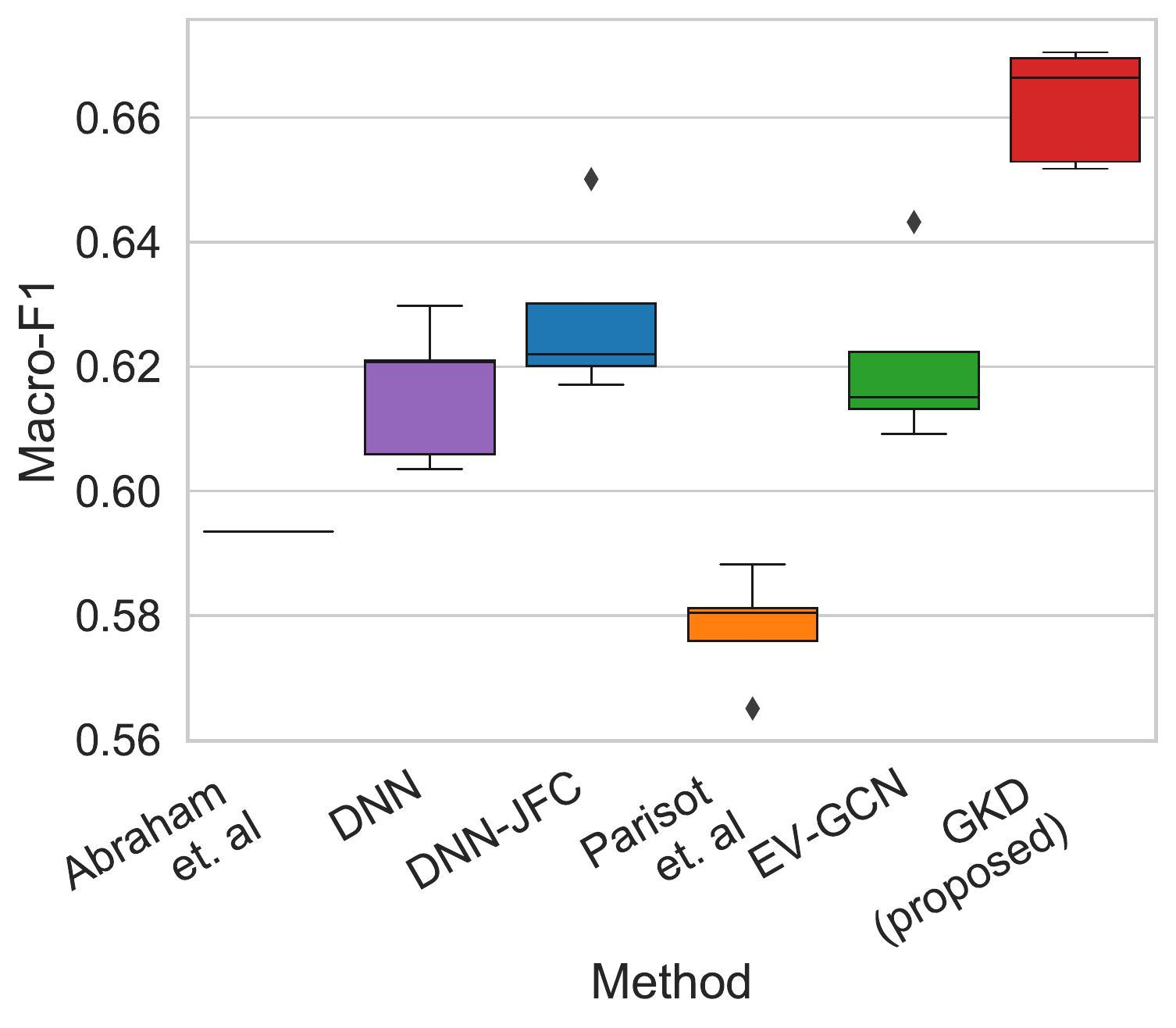}} 
    \subfloat[ABIDE AUC]{\label{fig:abide-auc}\includegraphics[width=0.33\textwidth]{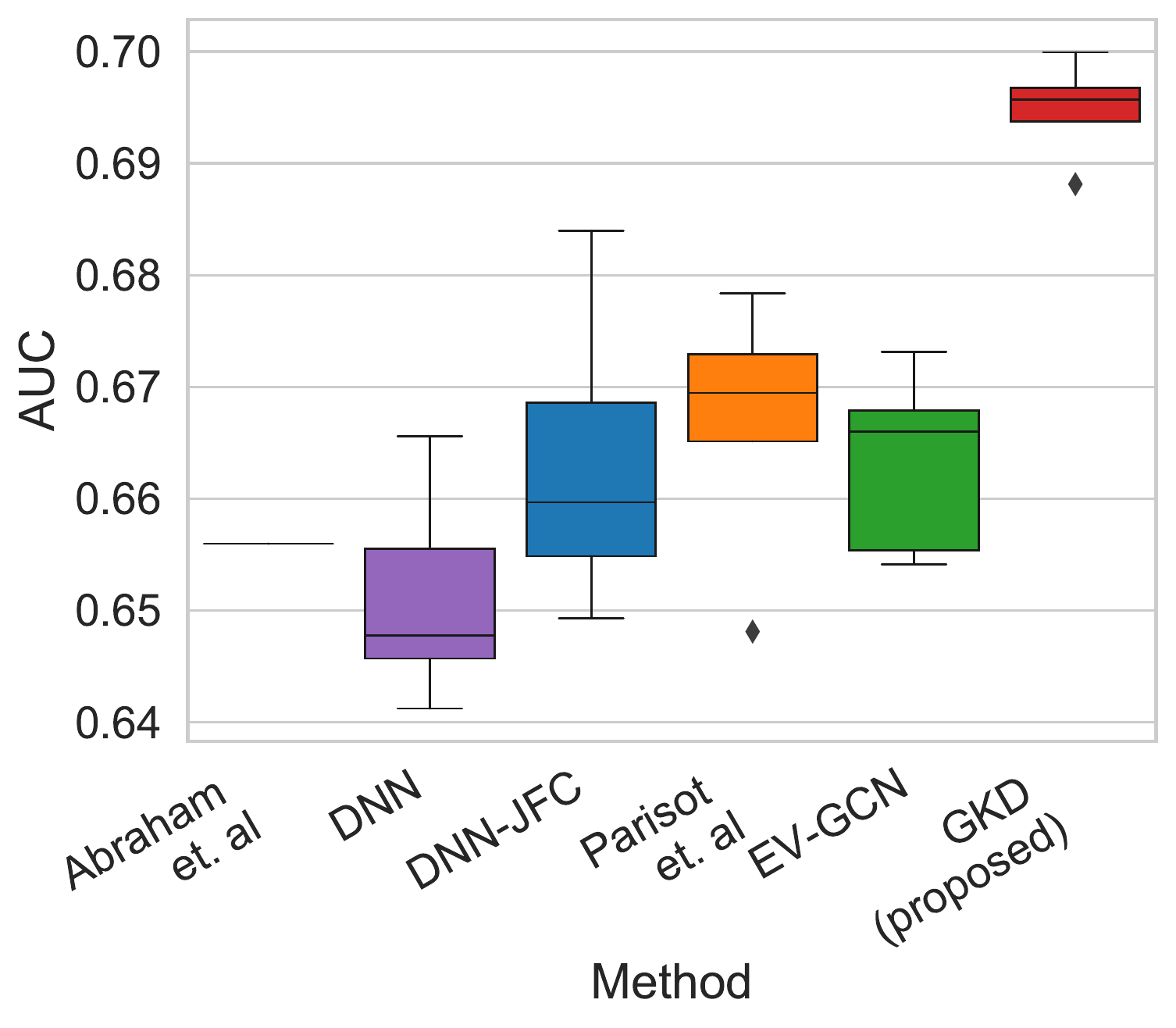}} 
  \end{center}
  \caption{Boxplot results of compared methods on the ABIDE dataset}
  \label{fig:abide}
}
\end{figure}

The Autism Brain Imaging Data Exchange (ABIDE) \cite{di2014autism,craddock2013neuro} database provides the neuroimaging (functional MRI) and phenotypic data
of $1112$ patients with binary labels indicating the presence of diagnosed Autism Spectrum Disorder (ASD).
To have a fair comparison with other methods, we choose the same $871$ patients comprising $468$ normal and $403$ ASD patients. We also follow the same pre-processing steps of Parisot et al. \cite{parisot2018disease} to select $2000$ features from fMRI for node features.
We use $55\%$, $10\%$, and $35\%$ for train, validation, and test split, respectively, and the labels are available for $40\%$ of training samples. 
For graph construction, first, we discard the phenotypic features that are only available for ASD patients, to prevent label leakage. The rest of the phenotypic features of patients are used for training a simple auto-encoder to both reconstruct the input via the decoder and classify the ASD state via a classifier using the latent low-dimensional representation of data (encoder's output). 
The auto-encoder is trained with the weighted sum of the mean squared errors as unsupervised (for labeled and unlabeled training samples) and cross-entropy as supervised (for labeled training samples) loss functions. \\
\textbf{Results and Analysis}: The boxplot of results are shown in \fig~\ref{fig:abide}. 
\method shows higher performance in all metrics. Improvement in accuracy and AUC indicates that \method's performance is not sensitive to the decision threshold and learns a more reliable latent representation than its competitors.
On the contrary, DNN shows appropriate accuracy, but it fails in AUC. This means that the embedding space of classes learned by DNN is not differentiable and selecting the right threshold is so effective.
On the other hand, it has been previously suggested that graphs are good choices for modeling multi-modal datasets \cite{parisot2018disease,huang2020edge}. We also provided the results of the methods when the graph is available during the inference time in \suptab~\ref{tab:abide-transductive} which also confirm that graph-based methods perform much better with the availability of the graph. However, in the current setting where the graph modality is not available, Parisot et al. and EV-GCN's performance drop by about $10\%$, indicating that the graph availability in inference time has an important role in their architecture.
\subsection{Alzheimer's Progression Prediction on the TADPOLE dataset}
\begin{figure}[!htb]
{
  \begin{center}
    \subfloat[TADPOLE Accuracy]{\label{fig:tadpole_binary-acc}\includegraphics[width=0.33\textwidth]{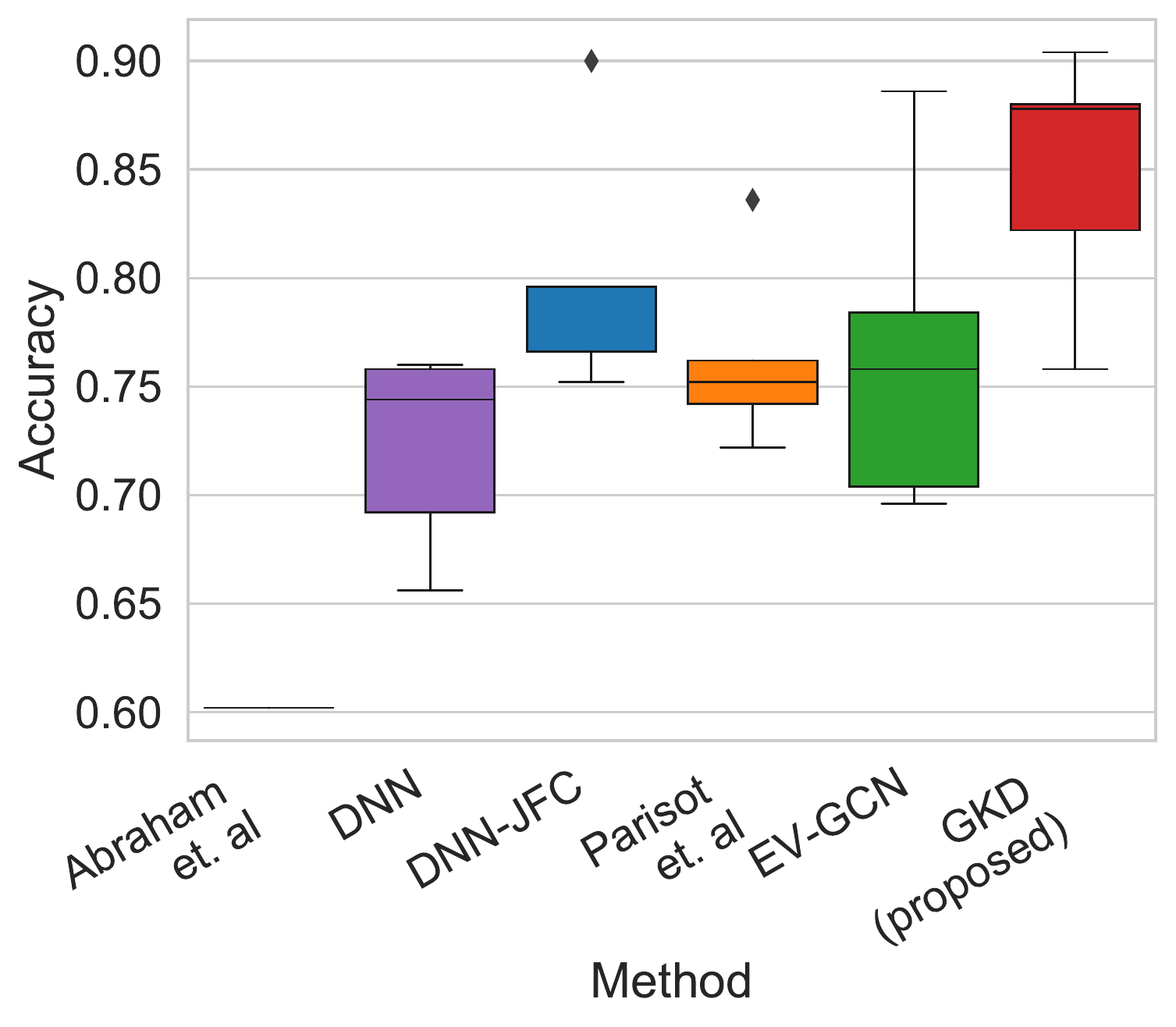}}
    \subfloat[TADPOLE Macro F1]{\label{fig:tadpole_binary-f1macro}\includegraphics[width=0.33\textwidth]{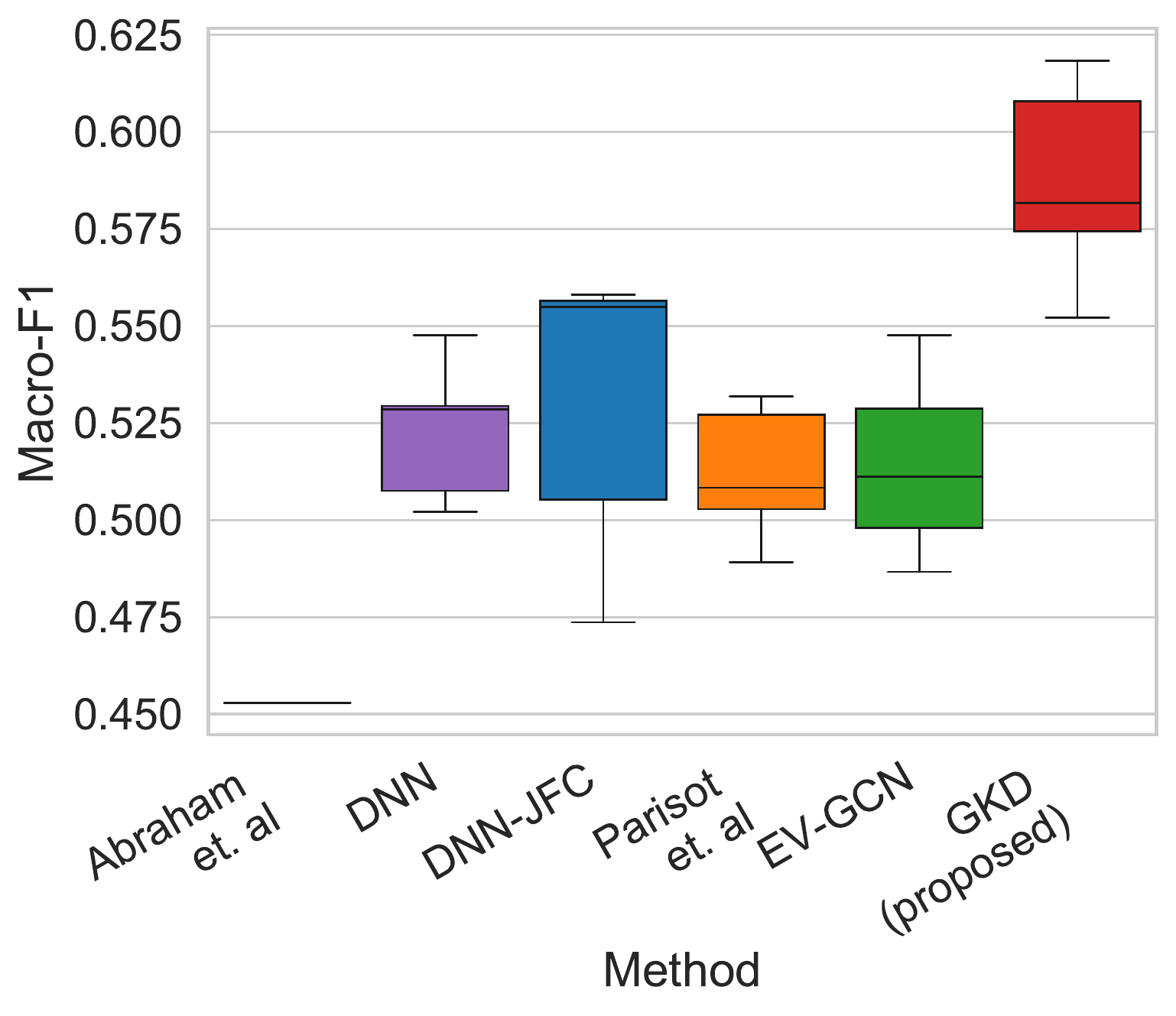}} 
    \subfloat[TADPOLE AUC]{\label{fig:tadpole_binary-auc}\includegraphics[width=0.33\textwidth]{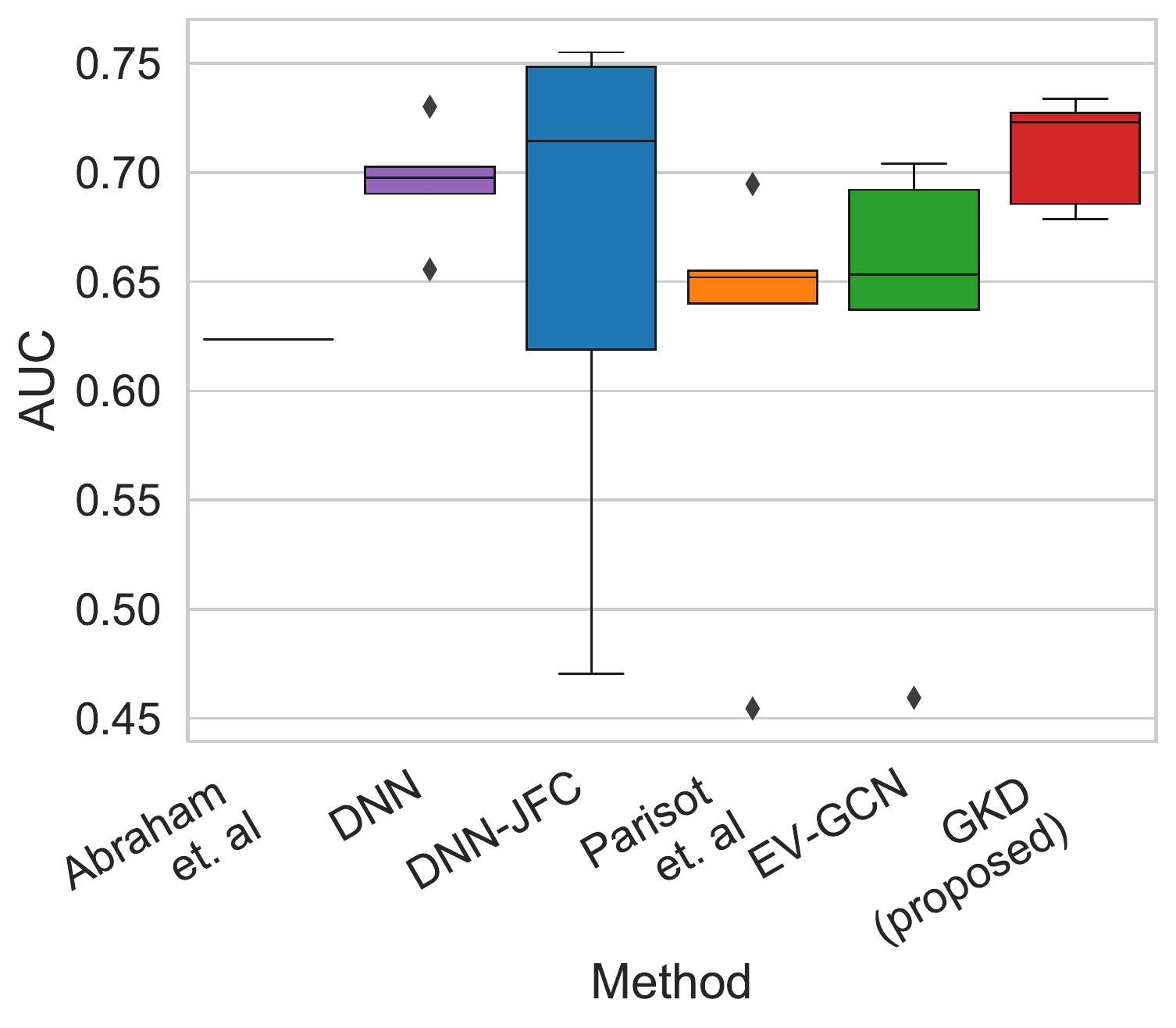}} 
  \end{center}
  \caption{Boxplot results of compared methods on the TADPOLE dataset}
  \label{fig:tadpole_binary}
}
\end{figure}

The Alzheimer's Disease Prediction Of Longitudinal Evolution (TAPOLE) \cite{marinescu2018tadpole} is a subset of Alzheimer’s Disease Neuroimaging Initiative (ADNI) data which is introduced in the TADPOLE challenge. The dataset has binary labels indicating if the patient's Alzheimer's status will be progressed in the next six months. The dataset contains $151$ instances with diagnosis or progression in Alzheimer's disease and $1878$ who keep their status. 
The $15$ biomarkers suggested by challenge organizers which are known to be informative in addition to the current status of patients are chosen as features. These biomarkers contains cognitive tests, MRI measures, PET measures, Cerebrospinal fluid (CSF) measures, APOE and age risk factors. 
We choose the biomarkers which are missing for more than $50\%$ of patients for constructing a graph based on their available values instead of imputing the missing ones. $A\beta$, $Tau$ and $pTau$ (CSF measures) and $FDG$ and $AV45$ (PET measures) are the sparse biomarkers. 
We follow the steps described in \cite{parisot2018disease} for graph construction. For each sparse biomarker, we connect every pair of nodes with the absolute distance less than a threshold. Then the union of constructed graphs are chosen as the final graph. More information about the graph modality is described in \suptab~\ref{tab:tadpole-info}. 
We use $65\%$, $10\%$, and $25\%$ for train, validation, and test split, respectively, and the labels are available for $10\%$ of training samples. \\
\textbf{Results and Analysis}: The results of Alzheimer's prediction are provided in \fig~\ref{fig:tadpole_binary}.
\method is superior to its competitors in all metrics. TADPOLE is a highly imbalanced dataset and enhancing accuracy and Macro F1 means efficiency of the proposed method in diagnosis of both classes. 
Graph-based methods (Parisot et al. and EV-GCN) show more stability and competitive results on the TADPOLE dataset. Hence, it can be concluded that the constructed graph is more effective in the ABIDE dataset and its absence at inference time leads to inadequate performance. 
Despite the ABIDE dataset, DNN-JFC demonstrates unstable results on the TADPOLE (even when both modalities are available during the test time, refer to \suptab~\ref{tab:tadpole-transductive}). 
Comparing DNN-JFC and DNN with graph-based methods indicates that using graph-modality features directly results in more instability than using them as graphs. 
\subsection{Synthetic Dataset}
\begin{figure}[!htb]
{
  \begin{center}
    \subfloat[Accuracy]{\label{fig:synthetic-acc}\includegraphics[width=0.33\textwidth]{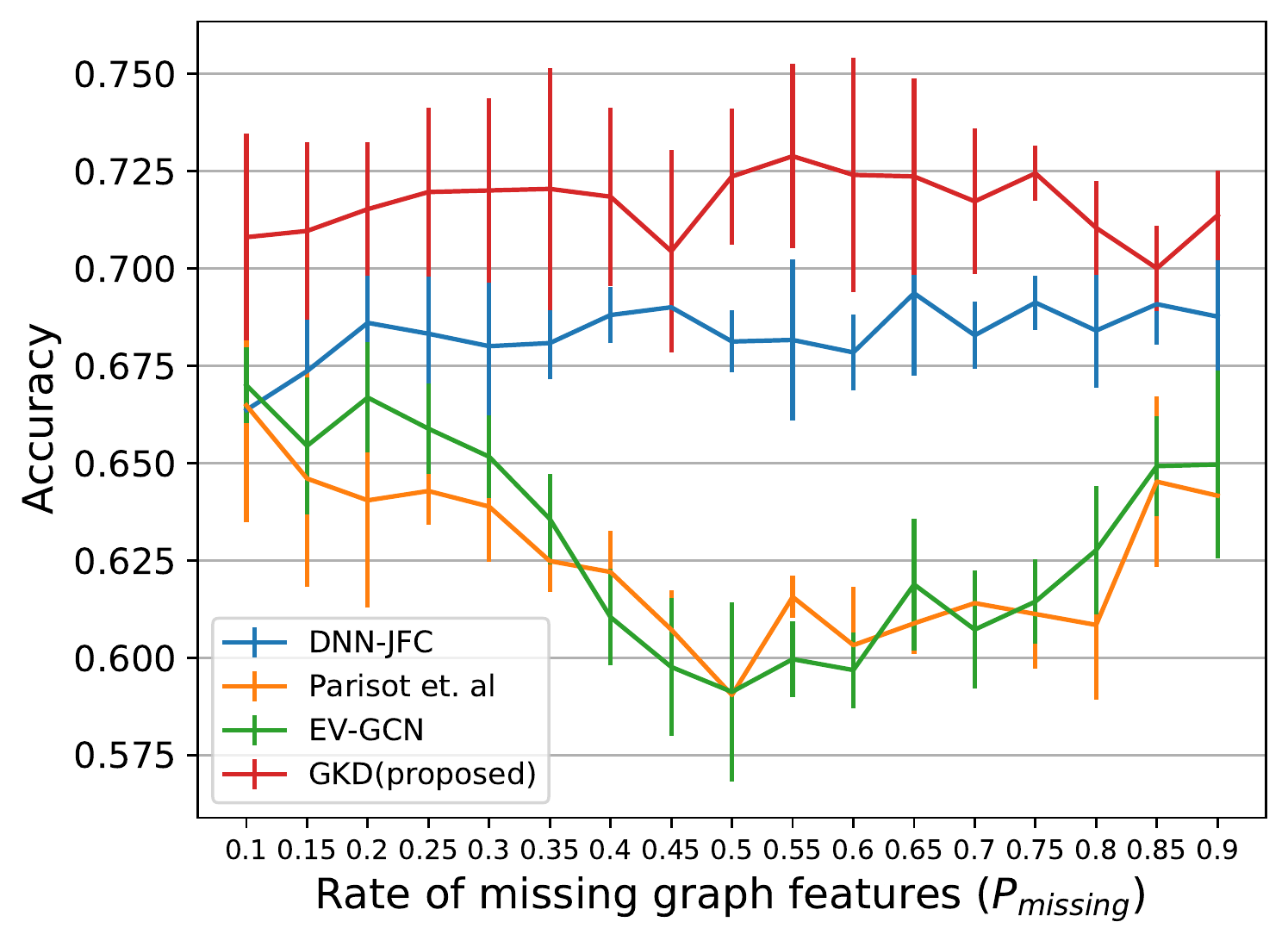}}
    \subfloat[Macro F1]{\label{fig:synthetic-f1macro}\includegraphics[width=0.33\textwidth]{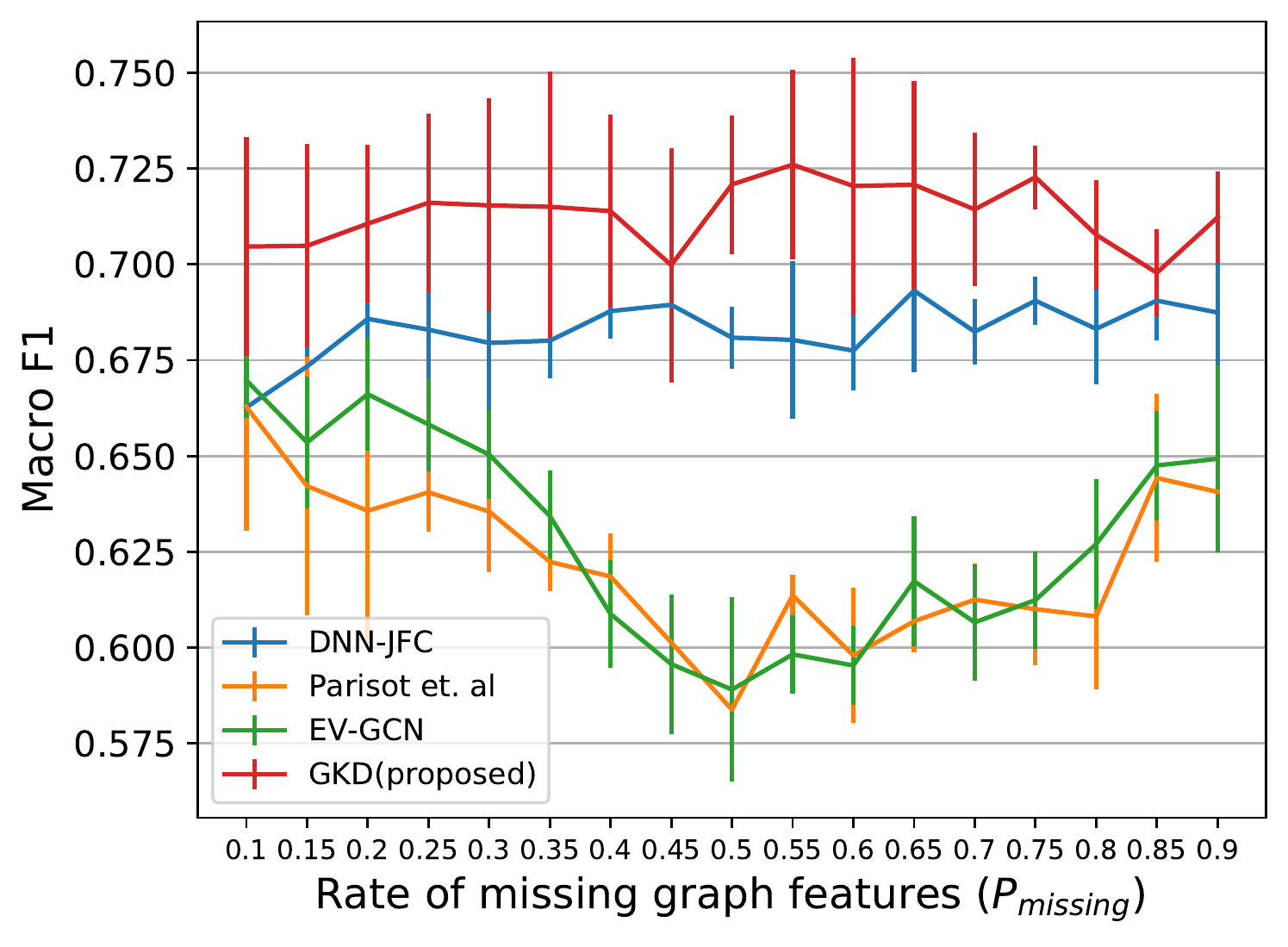}} 
    \subfloat[AUC]{\label{fig:synthetic-auc}\includegraphics[width=0.33\textwidth]{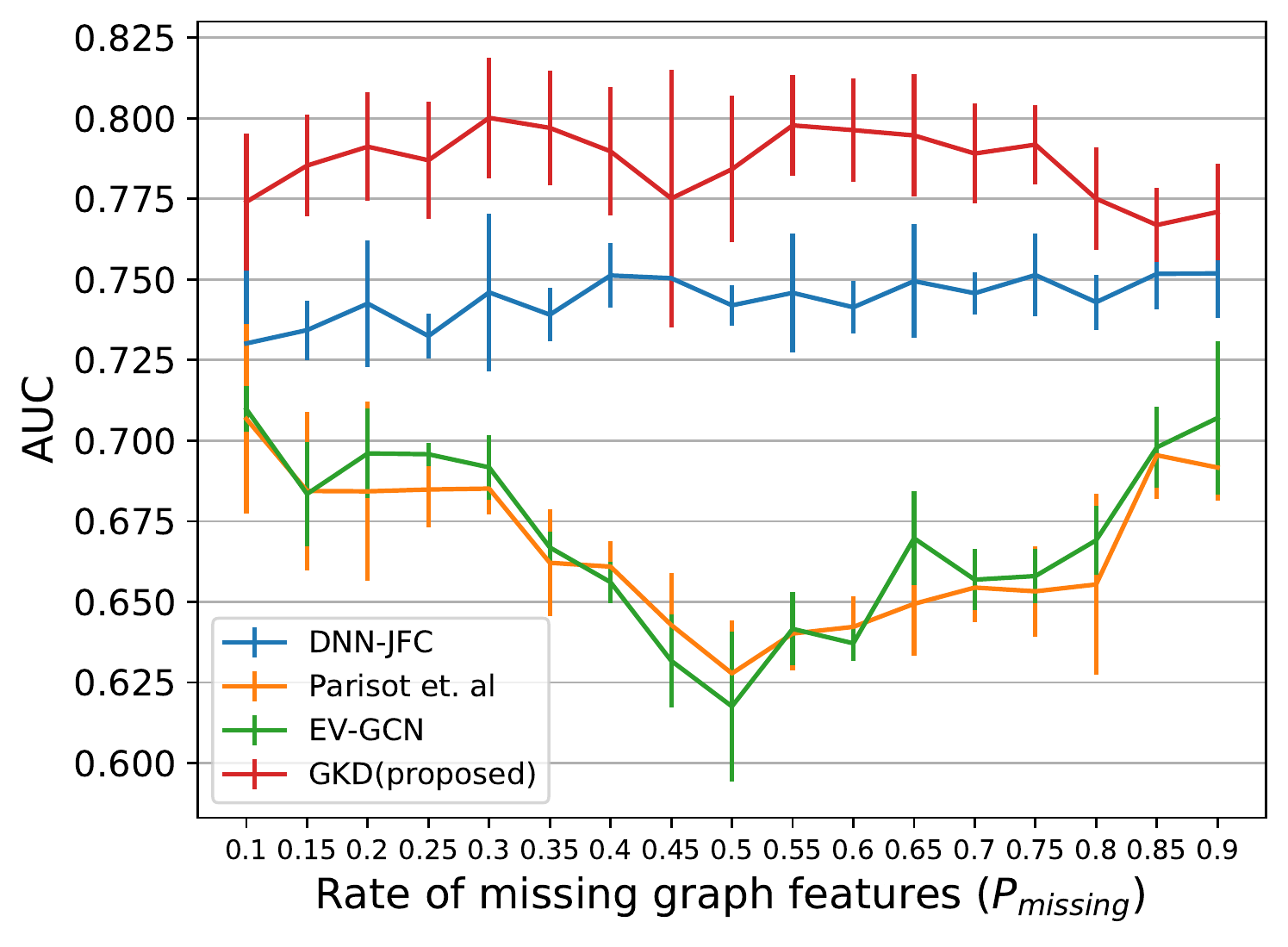}} 
  \end{center}
  \caption{Results of multi-modal methods by changing the rate of missing value in the graph-features at training}
  \label{fig:synthetic}
}
\end{figure}

To investigate the effect of graph modality in the training phase on the results of DNN-JFC, GCN, EV-GCN, and \method as multi-modal approaches, we create a set of synthetic datasets with different rates of missing values in the graph-modality. 
We generate $2000$ samples, distributed equally in two classes, with $128$-dimensional node features and $4$-dimensional graph-modality features. The algorithm of node-feature generation is adopted from \cite{guyon2003design}, which is based on creating class points normally distributed around vertices on a hypercube \cite{pedregosa2011scikit}.
The features of the graph modality are drawn from a $4$-dimensional standard normal distribution with mean $\mu_{c_1}$
and $\mu_{c_2}$
for each class, respectively.
Then we randomly remove the graph feature values by $P_{missing}$ rate and follow the steps of the TADPOLE graph construction. \\
\textbf{Results and Analysis}: The results of the experiment are illustrated in \fig~\ref{fig:synthetic}. 
This experiment also supports the claim that GCN-based architectures rely on the neighborhood in the graph, and inaccessibility to the graph deteriorates the efficiency of methods. 
As shown in \fig~\ref{fig:synthetic}, increasing the rate of missing values in training degrades the trained classifier (it also happens when the graph is available in the testing phase, refer to \supfig~\ref{fig:synthetic-false}), but with increasing the missing rate, the similarity between the training and testing data overcomes and the classifier improves. It should be noted that the shortcoming of GCN-based methods in comparison with DNN-JFC is another reason showing the weakness of GCN architecture in taking advantage of graph modality. 
It is also worth noting that GKD benefits from both features and graph and consistently outperforms in all metrics.  
\subsection{Discussion \& Conclusion}
This paper proposes a semi-supervised method to integrate distinct modalities (including imaging and non-imaging) into a unified predictive model that benefits from the graph information during the training phase and operates independently from graph modality during the inference time.

Unlike previous graph convolutional-based methods that rely on graph metadata, our \method method shows superior performance when the graph modality is missing among unseen samples. As a proof of concept, we evaluated our method for the task of diagnosing the Autism spectrum disorder on the ABIDE dataset and predicting Alzheimer's disease progress using the TADPOLE dataset. 
Extensive experimental results show that \method manifests a consistent enhancement in terms of accuracy, AUC,and Macro F1 compared to the SOTA methods. Additionally, the ability of the method on a set of semi-labeled synthetic datasets is examined.
We believe that this work opens the path to learn the classifier as it utilizes all the available information during the training time without being concerned about the availability of one or multiple modalities while diagnosing a patient's status. 

%
%
\bibliographystyle{splncs04}
\bibliography{ref}
%




\newpage
\section{Appendix}
\input{supplement_attached}
\end{document}

%% file: supplement_attached.tex
\begin{table}[!htb]
\centering
\caption{The table contains the features of TADPOLE dataset used for graph construction. The first row is the percentage of missing data in every feature. The second row shows the connectivity threshold used for connecting pair of nodes}
\small
\begin{tabular}{|c|c|c|c|c|c|}
\hline
Metric                                                             & $A\beta$ & $Tau$ & $pTau$ & $FDG$ & $AV45$ \\ \hline
\begin{tabular}[c]{@{}c@{}}Percentage of \\ Missing\end{tabular}      & 74.76    & 74.96 & 74.96  & 56.18 & 94.62  \\ \hline
\begin{tabular}[c]{@{}c@{}}Threshold for\\ Connecting\end{tabular} & 20       & 15    & 1.5    & 0.02  & 0.03   \\ \hline
\end{tabular}
\label{tab:tadpole-info}
\end{table}
\begin{table}[!htb]
\centering
\caption{Results corresponding to boxplot of Fig. 2 demonstrating performance on ABIDE dataset when the graph modality is not available in the testing phase}
\notsotiny
\begin{tabular}{|c|c|c|c|}
\hline
Method & Accuracy & F1 Macro & AUC \\ \hline \hline 
Abraham et. al & 0.607 ± 0.0 & 0.593 ± 0.0000 & 0.656 ± 0.0 \\ \hline
DNN & 0.629 ± 0.0065 & 0.616 ± 0.0099 & 0.651 ± 0.0086 \\ \hline
DNN-JFC & 0.632 ± 0.0185 & 0.628 ± 0.0119 & 0.663 ± 0.0121 \\ \hline
Parisot et. al & 0.6 ± 0.0130 & 0.578 ± 0.0076 & 0.667 ± 0.0103 \\ \hline
EV-GCN & 0.627 ± 0.0092 & 0.6210 ± 0.0121 & 0.663 ± 0.0074 \\ \hline
GKD(proposed) & \textbf{0.667 ± 0.0087} & \textbf{0.662 ± 0.0082} & \textbf{0.695 ± 0.0039} \\ \hline
\end
{tabular}
\label{tab:abide-inductive}
\end{table}
\begin{table}[!htb]
\centering
\caption{Results on TADPOLE dataset when the graph is available in the testing phase and testing samples are also accessible during the training}
\notsotiny
\begin{tabular}{|c|c|c|c|}
\hline
Method & Accuracy & F1 Macro & AUC \\ \hline \hline 
Abraham et. al & 0.607 ± 0.0 & 0.593 ± 0.0 & 0.656 ± 0.0 \\ \hline
DNN & 0.629 ± 0.0065 & 0.616 ± 0.0099 & 0.651 ± 0.0086 \\ \hline
DNN-JFC & 0.65 ± 0.0152 & 0.641 ± 0.0197 & 0.688 ± 0.0067 \\ \hline
Parisot et. al & 0.704 ± 0.0252 & 0.685 ± 0.0398 & 0.748 ± 0.0106 \\ \hline
EV-GCN & 0.701 ± 0.0309 & 0.69 ± 0.028 & 0.767 ± 0.0216 \\ \hline
GKD(proposed) & \textbf{0.714 ± 0.0076} & \textbf{0.707 ± 0.0091} & \textbf{0.769 ± 0.005} \\ \hline
\end
{tabular}
\label{tab:abide-transductive}
\end
{table}
\begin{table}[!htb]
\centering
\caption{Results corresponding to boxplot of Fig. 3 demonstrating performance on TADPOLE dataset when the graph modality is not available in the testing phase}
\notsotiny
\begin{tabular}{|c|c|c|c|}
\hline
Method & Accuracy & F1 Macro & AUC \\ \hline \hline 
Abraham et. al & 0.602 ± 0.0 & 0.453 ± 0.0 & 0.623 ± 0.0 \\ \hline
DNN & 0.722 ± 0.0412 & 0.523 ± 0.0164 & 0.695 ± 0.024 \\ \hline
DNN-JFC & 0.796 ± 0.0539 & 0.53 ± 0.0343 & 0.662 ± 0.1072 \\ \hline
Parisot et. al & 0.763 ± 0.0389 & 0.512 ± 0.0158 & 0.619 ± 0.0844 \\ \hline
EV-GCN & 0.766 ± 0.0686 & 0.514 ± 0.0217 & 0.629 ± 0.0883 \\ \hline
GKD(proposed) & \textbf{0.848 ± 0.0526} & \textbf{0.587 ± 0.0237} & \textbf{0.71 ± 0.0229} \\ \hline
\end
{tabular}
\label{tab:tadpole-inductive}
\end
{table}
\begin{table}[!htb]
\centering
\caption{Results on TADPOLE dataset when the graph is available in the testing phase and testing samples are also accessible during the training}
\notsotiny
\begin{tabular}{|c|c|c|c|}
\hline
Method & Accuracy & F1 Macro & AUC \\ \hline \hline 
Abraham et. al & 0.604 ± 0.0 & 0.454 ± 0.0 & 0.623 ± 0.0 \\ \hline
DNN & 0.764 ± 0.0882 & 0.509 ± 0.0174 & 0.669 ± 0.0621 \\ \hline
DNN-JFC & 0.797 ± 0.0868 & 0.508 ± 0.0272 & 0.641 ± 0.1126 \\ \hline
Parisot et. al & 0.866 ± 0.0157 & 0.559 ± 0.0408 & 0.555 ± 0.094 \\ \hline
EV-GCN & 0.864 ± 0.0125 & 0.583 ± 0.0128 & 0.619 ± 0.0512 \\ \hline
GKD(proposed) & \textbf{0.887 ± 0.012} & \textbf{0.585 ± 0.0308} & \textbf{0.738 ± 0.0314} \\ \hline
\end{tabular}
\label{tab:tadpole-transductive}
\end{table}
\begin{figure}[!htb]
{
  \begin{center}
    \subfloat[Accuracy]{\label{fig:synthetic-false-acc}\includegraphics[width=0.33\textwidth]{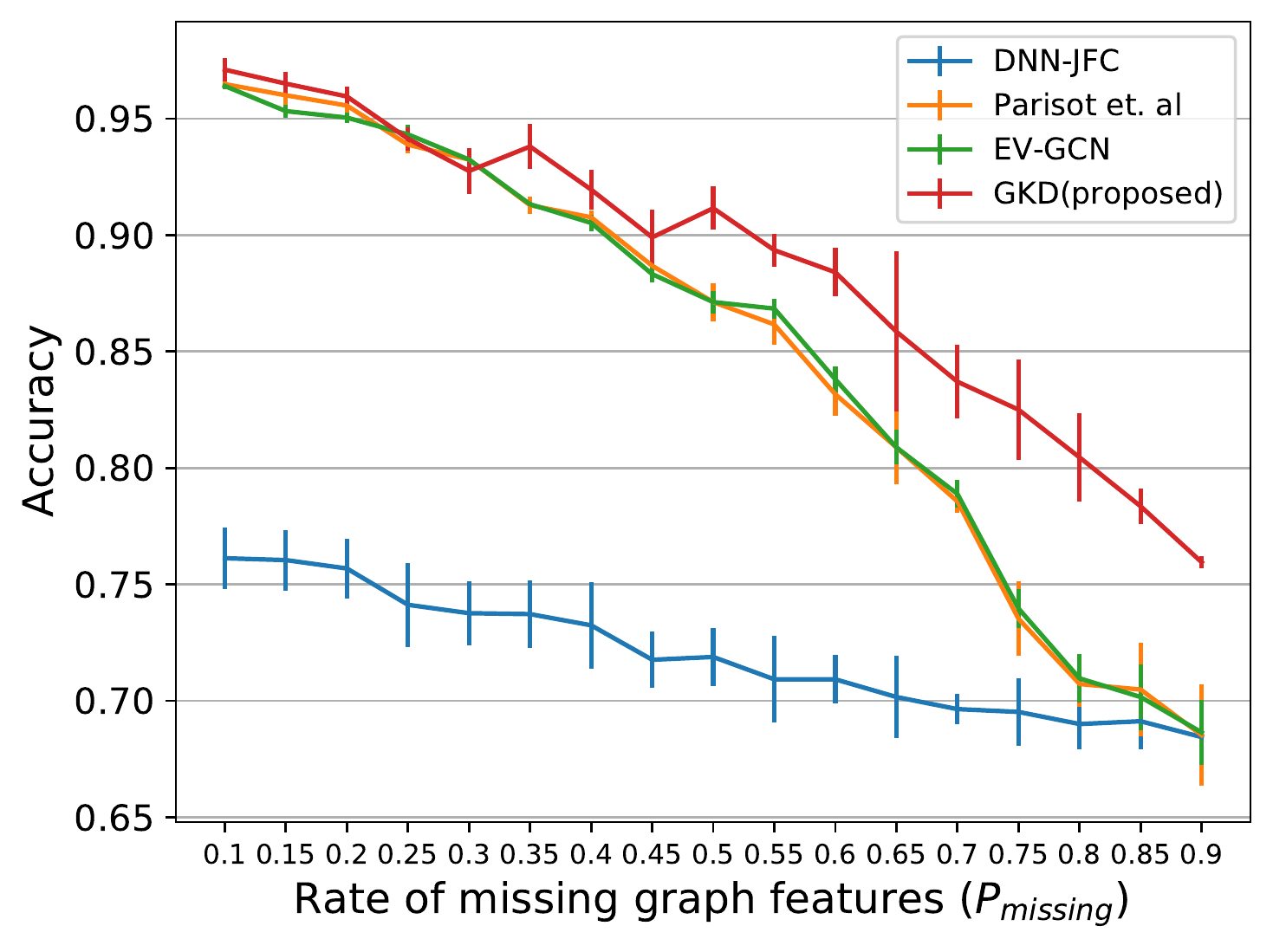}}
    \subfloat[Macro F1]{\label{fig:synthetic-false-f1macro}\includegraphics[width=0.33\textwidth]{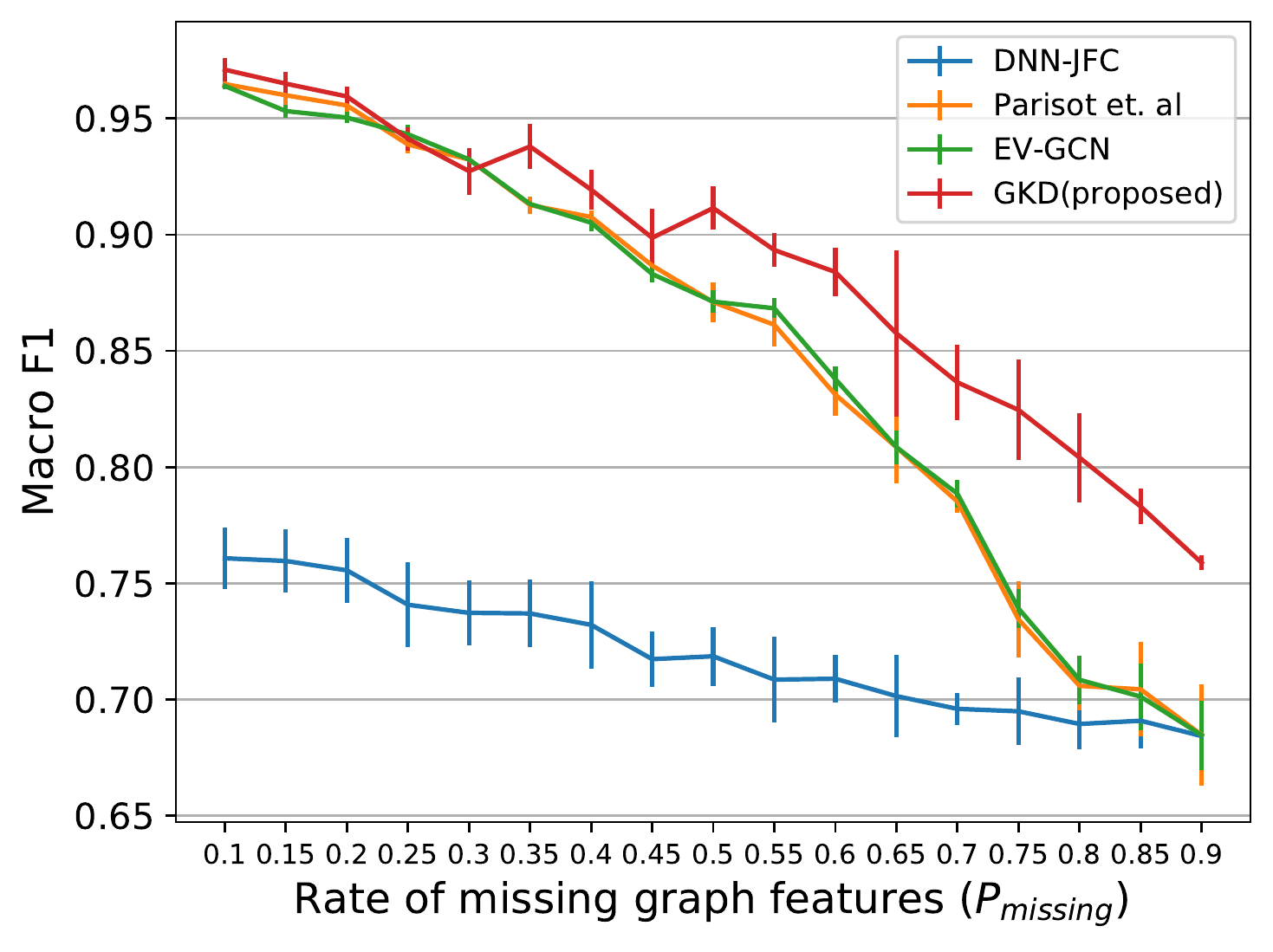}} 
    \subfloat[AUC]{\label{fig:synthetic-false-auc}\includegraphics[width=0.33\textwidth]{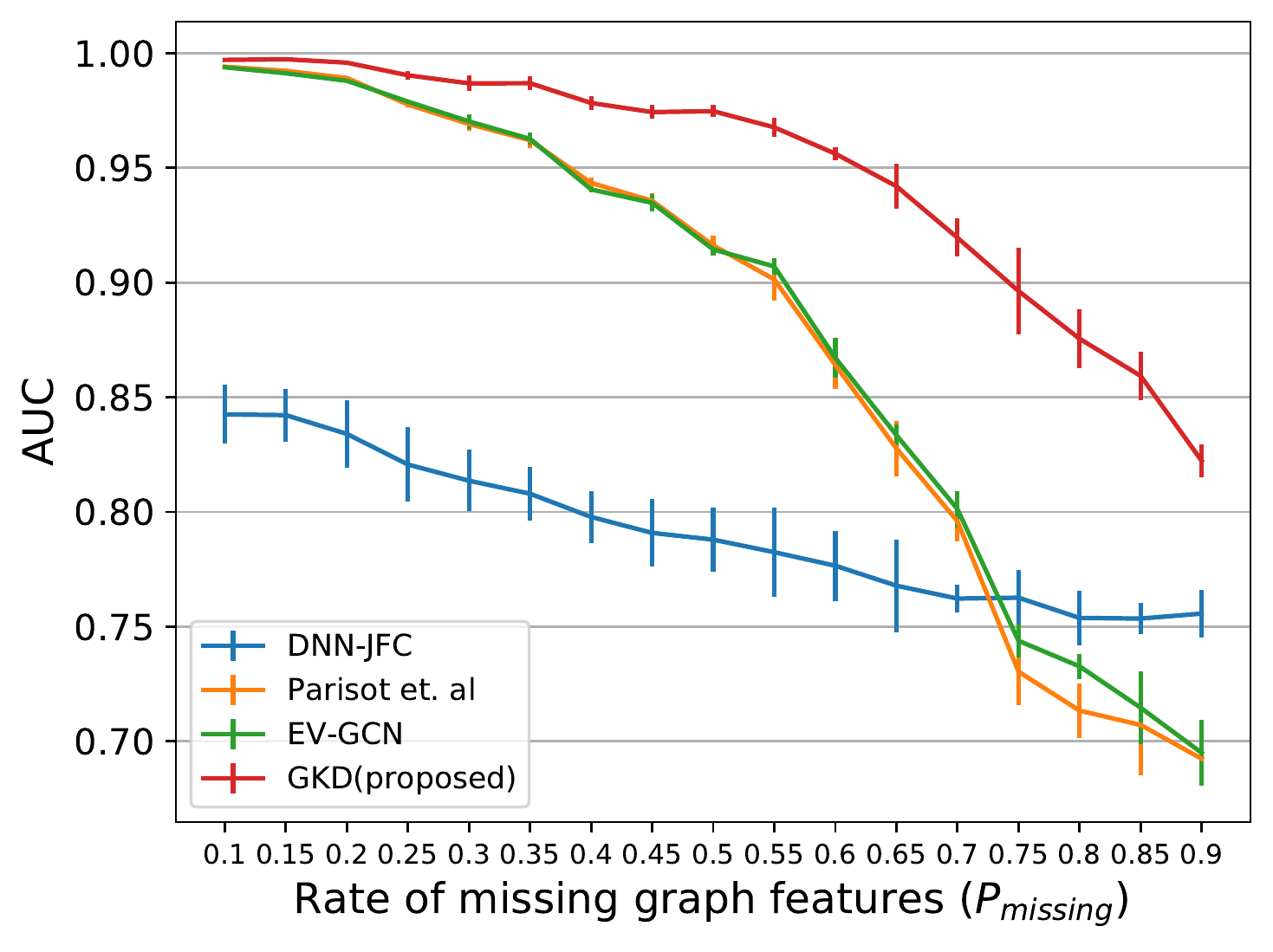}} 
  \end{center}
  \caption{Results of the multi-modal methods on synthetic datasets produced by changing the rate of missing values in the graph features. The graph is available in the testing phase and testing samples are also accessible during the training.}
  \label{fig:synthetic-false}
}
\end{figure}

%% file: main.bbl
\begin{thebibliography}{10}
\providecommand{\url}[1]{\texttt{#1}}
\providecommand{\urlprefix}{URL }
\providecommand{\doi}[1]{https://doi.org/#1}

\bibitem{abraham2017deriving}
Abraham, A., Milham, M.P., Di~Martino, A., Craddock, R.C., Samaras, D.,
  Thirion, B., Varoquaux, G.: Deriving reproducible biomarkers from multi-site
  resting-state data: An autism-based example. NeuroImage  \textbf{147},
  736--745 (2017)

\bibitem{abrol2019multimodal}
Abrol, A., Fu, Z., Du, Y., Calhoun, V.D.: Multimodal data fusion of deep
  learning and dynamic functional connectivity features to predict
  alzheimer’s disease progression. In: 2019 41st Annual International
  Conference of the IEEE Engineering in Medicine and Biology Society (EMBC).
  pp. 4409--4413. IEEE (2019)

\bibitem{bi2019effective}
Bi, X.a., Cai, R., Wang, Y., Liu, Y.: Effective diagnosis of alzheimer’s
  disease via multimodal fusion analysis framework. Frontiers in genetics
  \textbf{10}, ~976 (2019)

\bibitem{bucilu2006model}
Buciluǎ, C., Caruana, R., Niculescu-Mizil, A.: Model compression. In:
  Proceedings of the 12th ACM SIGKDD international conference on Knowledge
  discovery and data mining. pp. 535--541 (2006)

\bibitem{cai2019survey}
Cai, Q., Wang, H., Li, Z., Liu, X.: A survey on multimodal data-driven smart
  healthcare systems: approaches and applications. IEEE Access  \textbf{7},
  133583--133599 (2019)

\bibitem{craddock2013neuro}
Craddock, C., Benhajali, Y., Chu, C., Chouinard, F., Evans, A., Jakab, A.,
  Khundrakpam, B.S., Lewis, J.D., Li, Q., Milham, M., et~al.: The neuro bureau
  preprocessing initiative: open sharing of preprocessed neuroimaging data and
  derivatives. Frontiers in Neuroinformatics  \textbf{7} (2013)

\bibitem{defferrard2016convolutional}
Defferrard, M., Bresson, X., Vandergheynst, P.: Convolutional neural networks
  on graphs with fast localized spectral filtering. arXiv preprint
  arXiv:1606.09375  (2016)

\bibitem{di2014autism}
Di~Martino, A., Yan, C.G., Li, Q., Denio, E., Castellanos, F.X., Alaerts, K.,
  Anderson, J.S., Assaf, M., Bookheimer, S.Y., Dapretto, M., et~al.: The autism
  brain imaging data exchange: towards a large-scale evaluation of the
  intrinsic brain architecture in autism. Molecular psychiatry  \textbf{19}(6),
   659--667 (2014)

\bibitem{du2019zoom}
Du, H., Feng, J., Feng, M.: Zoom in to where it matters: a hierarchical graph
  based model for mammogram analysis. arXiv preprint arXiv:1912.07517  (2019)

\bibitem{ghorbani2021ragcn}
Ghorbani, M., Kazi, A., Baghshah, M.S., Rabiee, H.R., Navab, N.: Ra-gcn: Graph
  convolutional network for disease prediction problems with imbalanced data
  (2021)

\bibitem{guo2019deep}
Guo, Z., Li, X., Huang, H., Guo, N., Li, Q.: Deep learning-based image
  segmentation on multimodal medical imaging. IEEE Transactions on Radiation
  and Plasma Medical Sciences  \textbf{3}(2),  162--169 (2019)

\bibitem{guyon2003design}
Guyon, I.: Design of experiments of the nips 2003 variable selection benchmark.
  In: NIPS 2003 workshop on feature extraction and feature selection. vol.~253
  (2003)

\bibitem{hinton2015distilling}
Hinton, G., Vinyals, O., Dean, J.: Distilling the knowledge in a neural
  network. arXiv preprint arXiv:1503.02531  (2015)

\bibitem{huang2020fusion}
Huang, S.C., Pareek, A., Seyyedi, S., Banerjee, I., Lungren, M.P.: Fusion of
  medical imaging and electronic health records using deep learning: a
  systematic review and implementation guidelines. NPJ digital medicine
  \textbf{3}(1), ~1--9 (2020)

\bibitem{huang2020edge}
Huang, Y., Chung, A.C.: Edge-variational graph convolutional networks for
  uncertainty-aware disease prediction. In: International Conference on Medical
  Image Computing and Computer-Assisted Intervention. pp. 562--572. Springer
  (2020)

\bibitem{kazi2019inceptiongcn}
Kazi, A., Shekarforoush, S., Krishna, S.A., Burwinkel, H., Vivar, G.,
  Kort{\"u}m, K., Ahmadi, S.A., Albarqouni, S., Navab, N.: Inceptiongcn:
  receptive field aware graph convolutional network for disease prediction. In:
  International Conference on Information Processing in Medical Imaging. pp.
  73--85. Springer (2019)

\bibitem{kingma2014adam}
Kingma, D.P., Ba, J.: Adam: A method for stochastic optimization. arXiv
  preprint arXiv:1412.6980  (2014)

\bibitem{kipf2016semi}
Kipf, T.N., Welling, M.: Semi-supervised classification with graph
  convolutional networks. arXiv preprint arXiv:1609.02907  (2016)

\bibitem{lee2019predicting}
Lee, G., Nho, K., Kang, B., Sohn, K.A., Kim, D.: Predicting alzheimer’s
  disease progression using multi-modal deep learning approach. Scientific
  reports  \textbf{9}(1),  1--12 (2019)

\bibitem{li2020braingnn}
Li, X., Duncan, J.: Braingnn: Interpretable brain graph neural network for fmri
  analysis. bioRxiv  (2020)

\bibitem{liu2020identification}
Liu, J., Tan, G., Lan, W., Wang, J.: Identification of early mild cognitive
  impairment using multi-modal data and graph convolutional networks. BMC
  bioinformatics  \textbf{21}(6),  1--12 (2020)

\bibitem{marinescu2018tadpole}
Marinescu, R.V., Oxtoby, N.P., Young, A.L., Bron, E.E., Toga, A.W., Weiner,
  M.W., Barkhof, F., Fox, N.C., Klein, S., Alexander, D.C., et~al.: Tadpole
  challenge: Prediction of longitudinal evolution in alzheimer's disease. arXiv
  preprint arXiv:1805.03909  (2018)

\bibitem{parisot2018disease}
Parisot, S., Ktena, S.I., Ferrante, E., Lee, M., Guerrero, R., Glocker, B.,
  Rueckert, D.: Disease prediction using graph convolutional networks:
  Application to autism spectrum disorder and alzheimer’s disease. Medical
  image analysis  \textbf{48},  117--130 (2018)

\bibitem{parisot2017spectral}
Parisot, S., Ktena, S.I., Ferrante, E., Lee, M., Moreno, R.G., Glocker, B.,
  Rueckert, D.: Spectral graph convolutions for population-based disease
  prediction. In: International conference on medical image computing and
  computer-assisted intervention. pp. 177--185. Springer (2017)

\bibitem{pedregosa2011scikit}
Pedregosa, F., Varoquaux, G., Gramfort, A., Michel, V., Thirion, B., Grisel,
  O., Blondel, M., Prettenhofer, P., Weiss, R., Dubourg, V., et~al.:
  Scikit-learn: Machine learning in python. the Journal of machine Learning
  research  \textbf{12},  2825--2830 (2011)

\bibitem{venugopalan2021multimodal}
Venugopalan, J., Tong, L., Hassanzadeh, H.R., Wang, M.D.: Multimodal deep
  learning models for early detection of alzheimer’s disease stage.
  Scientific reports  \textbf{11}(1),  1--13 (2021)

\bibitem{xu2016multimodal}
Xu, T., Zhang, H., Huang, X., Zhang, S., Metaxas, D.N.: Multimodal deep
  learning for cervical dysplasia diagnosis. In: International conference on
  medical image computing and computer-assisted intervention. pp. 115--123.
  Springer (2016)

\bibitem{yang2019interpretable}
Yang, H., Li, X., Wu, Y., Li, S., Lu, S., Duncan, J.S., Gee, J.C., Gu, S.:
  Interpretable multimodality embedding of cerebral cortex using attention
  graph network for identifying bipolar disorder. In: International Conference
  on Medical Image Computing and Computer-Assisted Intervention. pp. 799--807.
  Springer (2019)

\bibitem{zhang2019graph}
Zhang, S., Tong, H., Xu, J., Maciejewski, R.: Graph convolutional networks: a
  comprehensive review. Computational Social Networks  \textbf{6}(1),  1--23
  (2019)

\end{thebibliography}
